\documentclass[11pt]{article}
\usepackage{graphicx}
\usepackage{authblk}
\usepackage[textwidth=4.5cm]{geometry}
\usepackage{blindtext}

\usepackage{indentfirst,csquotes}

\usepackage{amssymb,amsthm,amsmath}
\usepackage{xcolor,paralist,hyperref,titlesec,fancyhdr,etoolbox}
\usepackage{graphicx}
\usepackage{tabularx}
\usepackage{longtable}

\newcolumntype{C}{>{\centering\arraybackslash}X}
\usepackage{caption}
\usepackage{subcaption}
\usepackage{multirow}
\usepackage{colortbl} 
\usepackage{makecell} 

\usepackage{amssymb}
\usepackage[utf8]{inputenc}
\usepackage{graphicx}%
\usepackage{multirow}%
\usepackage{amsmath,amssymb,amsfonts}%
\usepackage{amsthm}%
\usepackage{mathrsfs}%
\usepackage[title]{appendix}%
\usepackage{manyfoot}%
\usepackage{booktabs}%
\usepackage{float}%
\usepackage{algorithm}%
\usepackage{algorithmicx}%
\usepackage{algpseudocode}%
\usepackage{listings}%
\usepackage{wrapfig}%
\usepackage[font=small,labelfont=bf,justification=justified, format=plain]{caption}%
\usepackage{subcaption}%
\usepackage[tableposition=top]{caption}%
\graphicspath{{figures/}}
\usepackage{lipsum}

\begin{document}
\title{Satellite-based feature extraction and multivariate time-series prediction of biotoxin contamination in shellfish}

\author[1]{S\'ergio Tavares}
\author[2]{Pedro R. Costa}
\author[3]{Ludwig Krippahl}
\author[1,4,5,*]{Marta B. Lopes}
\affil[1]{NOVA School of Science and Technology (NOVA FCT), Caparica, Portugal}
\affil[2]{IPMA - Instituto Português do Mar e da Atmosfera, 1495-165 Algés, Portugal} 
\affil[3]{NOVA Laboratory for Computer Science and Informatics (NOVA LINCS), NOVA FCT, Caparica, Portugal}
\affil[4]{Center for Mathematics and Applications (NOVA Math), NOVA FCT, Caparica, Portugal}
\affil[5]{NOVA Research \& Development Unit for Mechanical and Industrial Engineering (UNIDEMI), NOVA FCT, Caparica, Portugal}
\affil[*]{correspondence to: marta.lopes@fct.unl.pt}
\maketitle


\begin{abstract} 
Shellfish production constitutes an important sector for the economy of many Portuguese coastal regions, yet the challenge of shellfish biotoxin contamination poses both public health concerns and significant economic risks. Thus, predicting shellfish contamination levels holds great potential for enhancing production management and safeguarding public health. In our study, we utilize a dataset with years of Sentinel-3 satellite imagery for marine surveillance, along with shellfish biotoxin contamination data from various production areas along Portugal's western coastline, collected by Portuguese official control. Our goal is to evaluate the integration of satellite data in forecasting models for predicting toxin concentrations in shellfish given forecasting horizons up to four weeks, which implies extracting a small set of useful features and assessing their impact on the predictive models. We framed this challenge as a time-series forecasting problem, leveraging historical contamination levels and satellite images for designated areas. While contamination measurements occurred weekly, satellite images were accessible multiple times per week. Unsupervised feature extraction was performed using autoencoders able to handle non-valid pixels caused by factors like cloud cover, land, or anomalies. Finally, several Artificial Neural Networks models were applied to compare univariate (contamination only) and multivariate (contamination and satellite data) time-series forecasting. Our findings show that incorporating these features enhances predictions, especially beyond one week in lagoon production areas (RIAV) and for the 1-week and 2-week horizons in the L5B area (oceanic). The methodology shows the feasibility of integrating information from a high-dimensional data source like remote sensing without compromising the model's predictive ability.
\end{abstract}
\bigskip
{\textbf{Keywords:}
Harmful Algal Blooms, Shellfish contamination, Multivariate Time-series Forecasting, Satellite Imaging, Artificial Neural Networks, Autoencoders, Feature Extraction }
\bigskip

\section{Introduction}\label{sec1}

Shellfish production can be viewed as a sustainable alternative for protein production in a world with increasing demand for seafood and increasing challenges to feed a growing world population \cite{Suplicy2020,Olivier2020}. Shellfish, here seen as bivalve molluscs, are filter-feeding organisms that obtain their energy to grow from particles available in the water column, namely phytoplankton cells. Therefore, there is no need to add feed or any input of fresh/drinkable water. No fertilizers, pesticides, or antibiotics are used as well. Moreover, shellfish with their relevant filtration rates can have a positive impact on water transparency and even be cultivated combined with fish farms in an Integrated Multi-Trophic Aquaculture, which has been a strategy to implement more sustainable systems \cite{MacDonald2011,Park2021}.

However, from the thousands of phytoplankton species in the oceans that act as primary producers, a few are able to produce toxins, which under certain environmental conditions may bloom and reach a high number of cells in the seawater, leading to shellfish contamination. Although aquaculture, and particularly shellfish farming, has been a growing business sector, this activity is severely impacted by harmful algal blooms (HAB). 

HAB are recurrent natural phenomena. Depending on the phytoplankton species involved, notably high cell concentrations may emerge within the water column in response to favorable oceanographic conditions. In specific instances, these occurrences might even alter the color of the sea surface. As a consequence of HAB events, shellfish may accumulate toxins exceeding the regulatory and safety limits for human consumption. Under these circumstances, the food safety agencies temporarily close shellfish harvesting and prohibit the sale of shellfish in the market \cite{Braga2023}.

According to the EU regulations, the member states that produce shellfish must implement a monitoring program and an audited Official Control of their shellfish production. Each shellfish production area should be classified in terms of microbiological contaminants, and shellfish should be regularly tested, on a weekly basis, for HAB-toxins determination. Phytoplankton analysis of the seawater should be performed simultaneously \cite{EU2004}. Whenever the monitoring program reaches a positive result (i.e., toxin levels exceeding the regulatory limits) the precautionary closure of the shellfish harvesting is set in order to minimize the risk of acute intoxication.

The implemented strategy with precautionary closures to shellfish harvesting is well established and is considered an effective approach to minimize the health risk as the number of cases of shellfish poisoning has been highly reduced. However, this system is complex, requires intensive and expensive field sample collection and laboratory analysis, and only provides a reactive response to the problem. To minimize the economic impact on shellfish producers a more proactive system allowing anticipating the shellfish contamination is of great importance \cite{CRUZ_2022,Cruz2021}. In this sense, remote sensing based on satellite imagery has been a key research tool for early detection of algal blooms, including harmful events \cite{Ahn2006,Caballero2020,hill2020habnet}, either in marine or freshwater environments \cite{Mishra2019,Huo2022}, or even in polar ice sheets \cite{Wang2018}.   

Although there are some limitations for accurate satellite observations, such as unfavorable atmospheric conditions (clouds) or the presence of suspended material and coloured dissolved organic matter in coastal regions \cite{Caballero2020}, the vast potential of satellite imagery for the detection of algal blooms has increased deeply over the last years, leading the International Ocean Colour Coordinating Group \cite{IOCCG2021} of the Intergovernmental Oceanographic Commission of UNESCO to elaborate a report on ``Observation of Harmful Algal Blooms with Ocean Colour Radiometry'', where some strategies are pointed out to distinguish harmful blooms from a background of harmless phytoplankton, including dinoflagellate blooms associated with paralytic shellfish poisoning, blooms of the toxic diatom genus \textit{Pseudo-nitzschia}, blooms of the neurotoxin dinoflagellate \textit{Karenia brevis}, and cyanobacterial blooms \cite{IOCCG2021}. 

The role of artificial neural network (ANN) models has recently been evaluated for HAB prediction. In particular, multilayer perceptron (MLP) (e.g., \cite{ReFrHaYa97,Lee_et_al_2003,VELOSUAREZ2007361,Li_et_al_2014, Kim_et_al_2023}), convolutional neural networks (CNNs) \cite{hill2020habnet,Yussof2021,PYO2021117483} and long short-term memory (LSTM) networks \cite{LeeLee2018,Cho2018,ChoPark2019,Yussof2021,KIM2022118289}, with the latter being increasingly used to overcome the limitations of the former. Although still valid as a proxy for shellfish contamination and for complementing decision-making tools, HAB forecasts may not represent an optimal solution. There is no direct relationship between HAB events and shellfish contamination, and not every HAB event translates into contamination. Unlike for HAB forecasting, only a few studies attempted to forecast shellfish contamination, which arises from HABs and directly impacts industry and public safety management. Grasso et al. (2019) \cite{Grasso_2019} used an MLP with one hidden layer to predict closures to shellfish harvesting areas due to paralytic shellfish poisoning (PSP) toxins in blue mussels, one to ten weeks in advance. Cruz et al. (2022) \cite{CRUZ_2022} developed several MLP, CNN, and LSTM to predict contamination of mussels by diarrhetic shellfish poisoning (DSP) toxins up to 4 weeks in advance. Several biological and environmental time-series variables involved in HABs’ formation and shellfish contamination were used for model building, including chlorophyll a (chl-a), sea surface temperature (SST), toxic phytoplankton cell counts, and meteorologic variables (wind, atmospheric temperature, and rainfall).

Despite the acknowledged potential of satellite imagery to enhance the prediction of HABs and shellfish contamination, the predictive ability of multispectral or hyperspectral image data has rarely been investigated in ANN-based forecasting models, and exclusively aimed at HAB forecasting. Pyo et al. \cite{PYO2021117483} developed a CNN model with a convolutional block attention module to predict cyanobacterial cell concentrations, based on in situ data, simulated hydrodynamic features, and chl-a distribution maps obtained from airborne-generated hyperspectral images. The incorporation of chl-a maps in the attention module was shown to contribute to improving the model prediction accuracy at certain periods. CNNs and LSTMs were also successfully evaluated as part of a HAB detection system based on historical records and remote sensing-based datacubes obtained from MODIS sensors (e.g., SST, chl-a, and several spectral bands) to classify and discriminate between HAB and non-HAB events \cite{hill2020habnet}.

In this work, we evaluate the contribution of satellite data on a multivariate forecasting model to predict diarrhetic shellfish poisoning (DSP) in shellfish species across Portuguese production areas. We take as variables past values of contamination in shellfish and a time series of satellite images for the areas studied. Given the high dimensionality of the satellite data, we propose a methodology encompassing a prior step for dimensionality reduction to extract a small number of relevant satellite-based features using autoencoders. Following the successful implementation of ANN models for biotoxin contamination forecasting in mussels \cite{CRUZ_2022}, MLP, CNN, and LSTM architectures were trained on contamination data and the extracted remote sensing features from 2016 to 2020 and validated in 2021. Model performance improvements in predicting contamination events in 2022 were obtained on a case-by-case basis regarding the forecasting horizons (t+1 to t+4 weeks) and the different areas evaluated. Our approach shows the usefulness of incorporating available inexpensive information from a high-dimension data source like remote sensing, configuring a promising tool to improve available mechanisms used by shellfish farmers for production management. 

\section{Materials and methods}\label{sec2}

\subsection{Data}\label{subsec2.1}

\subsubsection{Contamination Data}\label{subsubsec2.1.2}

Data regarding DSP contamination of several shellfish species across the Portuguese shellfish production areas is available for public access on the website of the Portuguese Institute of the Sea and Atmosphere (IPMA) at \url{https://www.ipma.pt/pt/bivalves/zonas}. Data has been available since August 2014 and updated periodically, with measurements being performed on a weekly basis with some irregularity. The time frame selected for this data is the same considered for satellite images, between 26-04-2016 and 31-12-2022. 
The mussel species \textit{Mytilus galloprovincialis} is the most commonly monitored species, given its faster toxin accumulation rate, therefore being considered as an indicator species. However several other species are also monitored depending on their commercial importance in the corresponding production area. In order to increase the size of the dataset for data modeling purposes, all available species were considered for this work. 

\begin{figure}
\centering
   \includegraphics[width=0.35\textwidth]{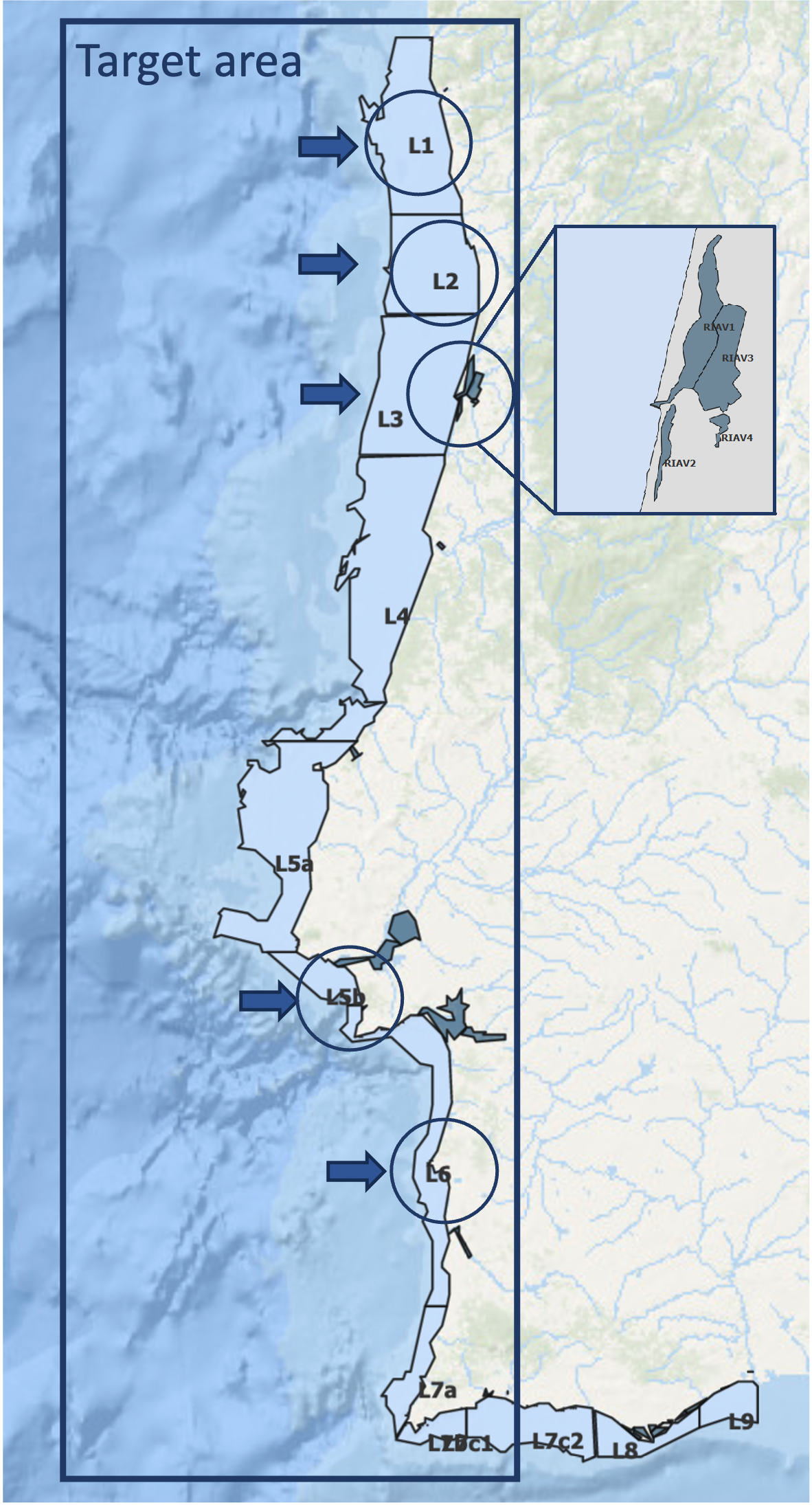}
  \caption{Selected locations (adapted from IPMA website at \url{https://www.ipma.pt/pt/bivalves/zonas/}). Areas RIAV1 to RIAV4 are lagoon areas adjacent to the L3 oceanic area.}
  \label{fig:locations}
\end{figure}

From the IPMA measurement sites on the western coast of Portugal, we selected for study the areas that featured a minimum amount of weekly data points (more than 200), while keeping a reasonable balance between contaminated and not contaminated samples (Table \ref{table:cont_dataset}). The selected areas were the following: L1, L2, L3, RIAV (RIAV1, RIAV2, RIAV3 and RIAV4), L5B, and L6 (Figure \ref{fig:locations}). A closer inspection of the contamination values reveals that contamination events are very likely to occur in consecutive weeks, and likewise for non-contamination. For this reason, we deal with missing values by assigning the previous value observed in the area, assuming it will maintain its current state until a future change.

\begin{table}[htbp]
\caption{Number of available weekly measurements for the selected areas, including the number of missing values and counts below or over contamination limits (Cont = contamination). The full dataset includes years 2016 to 2020, comprising a total of 349 weeks.}
\centering
\small
\begin{tabular}{@{}lccccc@{}}
\hline
\multicolumn{1}{l}{\textbf{Area}} &
  \multicolumn{1}{c}{\textbf{Measured}} &
  \multicolumn{1}{c}{\textbf{Missing}} &
  \multicolumn{1}{c}{\textbf{No Cont}} &
  \multicolumn{1}{c}{\textbf{Cont}} &
  \multicolumn{1}{c}{\textbf{Cont (\%)}} \\ \hline
L1                 & 216 & 133 & 129 & 87  & 25\% \\
L2                 & 318 & 31  & 201 & 117 & 34\% \\
L5B                & 250 & 99  & 151 & 99  & 28\% \\
L6                 & 261 & 88  & 208 & 53  & 15\% \\
RIAV1              & 327 & 22  & 142 & 185 & 53\% \\
RIAV2              & 346 & 3   & 165 & 181 & 52\% \\
RIAV3              & 338 & 11  & 238 & 100 & 29\% \\
RIAV4              & 215 & 134 & 141 & 74  & 21\% \\ \hline
\\
\end{tabular}
\label{table:cont_dataset}
\end{table}

\subsubsection{Satellite Data}\label{subsubsec2.1.2}

Satellite data was obtained from the Copernicus’ SENTINEL-3 mission, a part of the European Union’s space programme for earth observation, where several collections and levels of data processing are available for public access. The Ocean and Land Color Instrument (OLCI) provides useful information on the ocean and coastal areas with 2 processing levels available. Level 1B provides top of atmosphere radiances while level 2 further processes this data to water leaving reflectances and bio-optical variables (e.g., chlorophyll). Level 2 includes several atmospheric corrections as well as anomaly detection (e.g., masks for clouds and invalid pixels) hence being the most suitable data for our study. Satellite images come in multispectral data frames with a side of approximately 1440 km and when at full resolution each pixel covers an area of 300$\times$300m.
Data was selected from two collections: EO:EUM:DAT:0407 and EO:EUM:DAT:0556. The first one is the operational collection which allows access to the latest data available (typically one year). For older archive information one must access the reprocessed collection, the second one used. Several years of full-resolution multispectral images were gathered, starting on the first available date (26/04/2016) until the end of 2022, roughly 6 and a half years. The data was obtained using the \texttt{eumdac} Python library, made available by Europe’s meteorological satellite agency (EUMETSAT). It provides simple access to EUMETSAT data from a variety of satellite missions, with several useful command-line utilities for data search, download, and processing \cite{EUMDAC}.

\begin{figure}
\centering
    \begin{subfigure}[b]{0.68\textwidth}
    \centering
    \includegraphics[width=8cm]{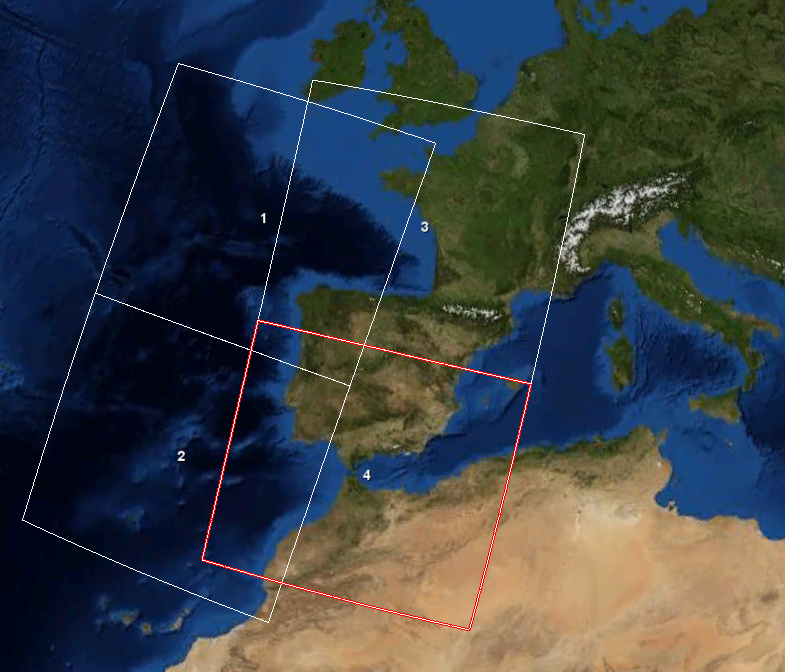}
    \caption{Example of satellite frames.}
    \label{fig:snap}
    \end{subfigure}
    \begin{subfigure}[b]{0.30\textwidth}
    \centering
    \includegraphics[width=3.3cm]{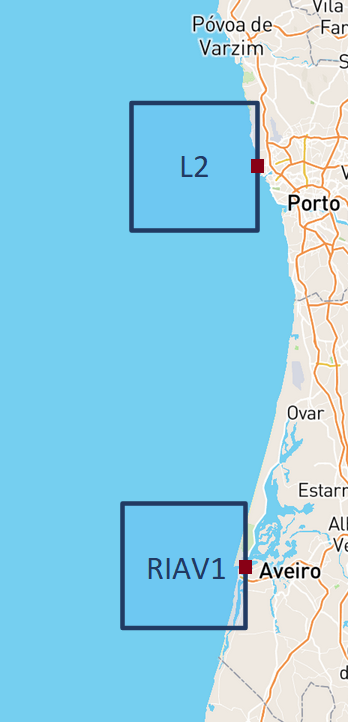}
    \caption{Image selection.}
    \label{fig:example_patches}
    \end{subfigure}
\caption{a) Satellite frame exploration on 25/04/2021 performed with SNAP software \cite{SNAP}. Frames 1 and 2 refer to satellite 3-A, while the remaining come from satellite 3-B (notice that the orbits and overlaps of both satellites differ); b) Example of final image selection areas for L2 and RIAV1 (red dots correspond to IPMA measurement sites and blue squares indicate the $64\times$64 extracted images.}
\end{figure}

After obtaining the frames from both collections, all daily information was aggregated. Ideally, full information would be retrieved over our target area (Western Portugal). However, the amount of coverage in each frame varies with satellite orbit (Figure \ref{fig:snap}). The mission comprises two satellites, A and B, which may each hold none, 1 or 2 contiguous frames overlapping the target area. Obtaining the largest possible coverage of the target area for each particular date requires a careful frame selection. Frames are gathered from the satellite covering more area in each date, using the strategy as follows: i) remove any frames with less than 1.5\% of coverage; ii) remove all frames from a given satellite if their combined coverage is under 20\% of the target area; iii) select remaining frames from the same satellite with most total coverage (e.g., two frames from satellite 3-B), with contiguous frames being merged.

As the goal was to obtain a set of images related to the contamination measurement sites from IPMA, further processing was required. To create these images we considered the coordinates of the measurement site to be in the center of the eastern edge of the image because we are considering sampling points along the western coast of Portugal (Figure \ref{fig:example_patches}). For each location, we select the closest point in the frame as reference to extract a 64x64 pixel multispectral image, corresponding to an area of 19,2$\times$19,2km. We expect this image size to be informative for our models and small enough to allow effective feature extraction. Image pixels referring to clouds, land, or several kinds of anomalies should be filtered out, so it was necessary to compute flags for invalid pixels for each image channel, based on the recommendations provided in the sentinel-3 manual \cite{EUMETSAT}. Images with 10\% or less valid pixels were discarded from the dataset.
Images were extracted directly from the frame without resampling, which may lead to slight orientation differences depending on the orbit of the satellite on a particular date.

\subsection{Time Series}\label{subsec2.3}

To compare the performance of models with or without satellite information, two different inputs were evaluated: a univariate case, which includes only information from contamination, and a multivariate case, where satellite features were also considered. 
Each time step corresponds to the maximum weekly value of contamination. Mussels show higher values of contamination overall, hence being selected in most cases when measurements from several species are present.
Features extracted from satellite data for the multivariate model correspond to the same or closest previous date to the measurement date. Each feature, as well as the contamination values, were rescaled separately to the [0, 1] range using min-max normalization.
Forecasts took as input the last 12 time steps, corresponding to the last three months, and targeted the upcoming 4-time steps. This input value was selected based on previous work performed on the same dataset, where an input of 12-time steps was found to provide the best results \cite{art_cruz2022forecasting}.

\subsection{Artificial Neural Network models}\label{subsec2.5}

The forecasting models developed are based on artificial neural networks (ANNs) commonly used for time series that proved successful in forecasting shellfish contamination \cite{art_cruz2022forecasting}. Several models were selected and compared, i.e., Multilayer Perceptron (MLP), Convolutional Neural Network (CNN), and Long Short-Term Memory (LSTM).

MLPs were the first ANN used for time series forecasting \cite{liu2020hybrid}. They consist of layers of neurons connected with weighted links, including an input layer, an output layer, and hidden layers. Neurons perform calculations and adjust their weights through training to optimize the network's response, typically using the back-propagation algorithm \cite{rumelhart1986learning, heaton2018ian}.

A CNN is a specialized type of ANN designed for processing grid-like data, such as images and time series. Instead of matrix multiplication, CNNs perform convolutions that involve sliding a weight matrix (kernel or filter) of a given size over the input data to create feature maps. The main idea is for each layer to learn a weight matrix that extracts important features from the input \cite{heaton2018ian}.

LSTM networks \cite{hochreiter1997long}, a type of Recurrent Neural Network, excel in capturing long data patterns by preserving gradient information over time, proving to be a successful approach in machine learning \cite{heaton2018ian}. In an LSTM, there are two distinct flows of information. At each time step, it processes the current input element and receives the hidden state and cell state from the previous time step. The hidden state results from non-linear transformations of the input, while the cell state is a function of linear transformations \cite{hewamalage2021recurrent}. This property allows the cell state to better preserve gradients and to handle both long and short time dependencies, hence this model is expected to perform best overall in our tests.

Details on the ANN architectures developed are presented in Table \ref{tab:ann_models}, respecting the number of batches, optimizer (e.g., Adam and RMSProp), kernel size, activation functions (e.g., sigmoid, ELU, ReLU, leaky\_ReLU) and layer dimensions tested. The ANN models were built and trained with the Keras library of the TensorFlow machine learning platform.

\begin{table}[htbp]
\centering
\begin{minipage}{\textwidth}
\caption{Description of the models and parameters tested when optimizing the time-series forecasts.}
\centering
\resizebox{1\textwidth}{!}{\begin{tabular}{@{}llc@{}}
\hline
\multicolumn{1}{c}{\textbf{Model}} & \multicolumn{1}{c}{\textbf{Description}} & \textbf{Layer Dimensions} \\ \hline
MLP  & Batch 8, Adam optimizer, ReLU activation, final sigmoid        & 12,24,36,48,60 \\
CNN  & Batch 8, Adam optimizer, kernel\_size 1,  leaky\_ReLU activation, final sigmoid & 12,24,36,48,60 \\
LSTM & Stateful (batch 1), RMSProp optimizer, final activation ELU    & 12,24,36,48,60\\
\hline
\end{tabular}}
\label{tab:ann_models}
 \end{minipage}
\end{table}

\subsection{Model Evaluation Metrics}\label{subsec2.4}

The problem under study was treated as a regression problem and evaluated based on standard regression metrics. Both the Mean Absolute Error (MAE) and the Root Mean Squared Error (RMSE) were considered, given by

\begin{equation}
\begin{aligned}
\text{MAE}&= \frac{1}{n} \sum_{i=1}^{n} |y_i - \widehat{y}_i| \qquad , \qquad \text{RMSE}&=\sqrt{\frac{1}{n} \sum_{i=1}^n\left(y_i-\hat{y}_i\right)^2}, \nonumber
\end{aligned}
\end{equation}

\noindent with $y_i =$ representing the true value and $\widehat{y}_i =$ the predicted value.

MAE was used as a loss function for training since it is easier to interpret in the context of this work and is less sensitive to outliers.

The classification above or below contamination limits is essential for regulating production areas, therefore classification accuracy was also computed, indicating the percentage of correctly classified cases, as follows

\begin{equation}
\text{Accuracy} = \frac{TP+TN}{\text{Total Cases}}, \nonumber
\end{equation}
\vspace{0pt}

\noindent with $TP$ representing true positives and $TN$ the true negatives cases.

\subsection{Feature Extraction}\label{subsec2.2}

Due to the inherently high dimensionality of the image data, a crucial step of the methodology proposed was to extract a small set of features retaining relevant information from the data. We propose to accomplish this in an unsupervised manner using autoencoders that are capable of ignoring all non-valid pixels. For this purpose, we focused on convolutional autoencoders, which are better equipped to deal with visual data, and evaluated several architectures. As the information extracted is expected to be specific to the geographical location, a different autoencoder was trained for each location.
The multispectral images included 16 water reflectance bands corresponding to different wavelengths and several bio-optical products. Previous work applied feature classification using a Random Forest to imaging products from NASA's MODIS satellite, estimating the importance of bands and products available for HAB forecasting \cite{hill2020habnet}. Following the authors' conclusions, and given available products are similar to the ones from SENTINEL-3 satellites, we selected CHL\_OC4ME and CHL\_NN (chlorophyll concentration estimation products) and Photosynthetically Active Radiation (PAR) as the most promising bio-optical products for our experiments.

\begin{figure}[htbp]
  \resizebox{1 \textwidth}{!}{   
  \centering
  \includegraphics[scale=1]{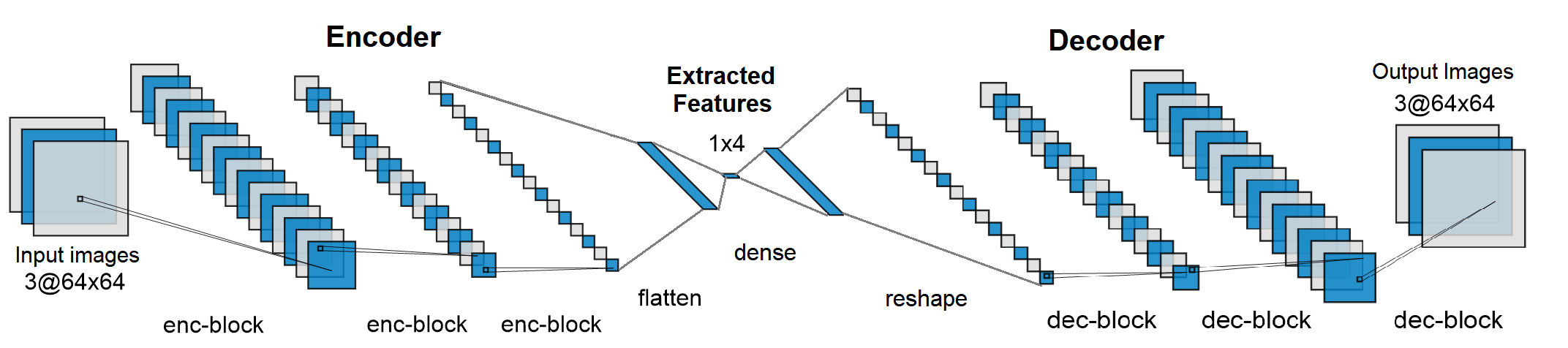}}
  \caption{3-Layer architecture used for the convolutional autoencoders.}
  \label{fig:autoencoder}
\end{figure}

Figure \ref{fig:autoencoder} shows the chosen architecture design for the convolutional autoencoders. Encoding and decoding blocks follow one of two patterns: conv-pool or conv-conv-pool (Table \ref{tab:conv_schemes}). Initial testing showed better performance for models with 3 layers. Parameters optimized include kernel size and number of filters (Table \ref{table:ker_fil}). 

\begin{table}[htbp]
\caption{Conv-Pool and Conv-Conv-Pool schemes, with the layers used for encoder and decoder blocks. Values tested for kernel sizes and number of filters are displayed  in  Table \ref{table:ker_fil}.}
\label{tab:conv_schemes}
\centering
\small
\begin{tabular}{@{}lclcc@{}}
\hline
\multicolumn{1}{l}{\textbf{Architecture}} &
  \textbf{Block} &
  \multicolumn{1}{c}{\textbf{Layers}} &
  \multicolumn{1}{c}{\textbf{Kernel}} &
  \multicolumn{1}{c}{\textbf{Filters}} \\ \hline
\multirow{6}{*}{Conv-Pool}      & \multirow{3}{*}{Encoder}                     & Conv2D             & K   & N  \\
                                &                                              & MaxPool            & 2$\times$2 & -  \\
                                &                                              & BatchNormalization & -   & -  \\ \cmidrule(l){2-5} 
                                & \multirow{3}{*}{Decoder}                     & UpSampling2D       & 2$\times$2 & 16 \\
                                &                                              & Conv2D             & K   & N  \\
                                &                                              & BatchNormalization & -   & -  \\ \hline
\multirow{8}{*}{Conv-Conv-Pool} & \multirow{4}{*}{Encoder}                     & Conv2D             & K   & N  \\
                                &                                              & Conv2D             & K   & N  \\
                                &                                              & MaxPool            & 2$\times$2 & -  \\
                                &                                              & BatchNormalization & -   & -  \\ \cmidrule(l){2-5} 
                                & \multicolumn{1}{l}{\multirow{4}{*}{Decoder}} & UpSampling2D       & 2$\times$2 & 16 \\
                                & \multicolumn{1}{l}{}                         & Conv2D             & K   & N  \\
                                & \multicolumn{1}{l}{}                         & Conv2D             & K   & N  \\
                                & \multicolumn{1}{l}{}                         & BatchNormalization & -   & -  \\ \hline
\end{tabular}
\end{table}

Model selection was performed based on the MSE, disregarding invalid pixels. A different model was selected for each area.

\begin{table}[htbp]
\captionsetup{width=0.95\textwidth}
\caption{Parameters tested when optimizing autoencoders, resulting in 16 possible combinations. The best parameter combinations are highlighted.}
\label{table:ker_fil}
\setlength{\tabcolsep}{12pt}
\centering
\small
\begin{tabular}{cc}
\hline
\textbf{Kernel Sizes (K)} & \textbf{Filter Numbers (N)} \\ \hline
(3$\times$3, 3$\times$3, 3$\times$3)       & (16, 16, 16)            \\
(5$\times$5, 5$\times$5, 5$\times$5)       & \textbf{(32, 32, 32)}            \\
\textbf{(3$\times$3, 5$\times$5, 5$\times$5)}       & (16, 32, 64)            \\
\textbf{(3$\times$3, 5$\times$5, 7$\times$7)}       & \textbf{(32, 64, 128)}           \\ \hline 
\\
\end{tabular}
\end{table}

Autoencoders were trained with Adam optimizer, learning rate=0.001, batch size=16, and 1000 epochs with early stopping (patience=100). Similarly to the ANN models, autoencoders were built and trained with the Keras library of the TensorFlow machine learning platform and tensorflow-gpu was used for increased performance. Data from 2016-2020 was used for training and 2021-2022 for validation. Both architectures were relevant when selecting the best models, (3$\times$3, 5$\times$5, 5$\times$5) and (3$\times$3, 5$\times$5, 7$\times$7) for kernel size and (32, 32, 32), (32, 64, 128) for filter number being the best combinations for these parameters amongst all areas. The best models selected were retrained using the full dataset before extracting features. Only four features were extracted in the middle layer, as they proved sufficient to improve results in several areas.

\section{Results and discussion}\label{sec3}

This section provides a summary of the findings of our experiments, comparing the performance of the different networks across the several shellfish production areas studied and regarding the various forecasting horizons and metrics considered.

The results obtained for all regions studied are displayed in Table \ref{table:results}, referring to the metrics evaluated on the test set (2022). The univariate case respects to using only past values of contamination for forecasting, whereas the multivariate case also incorporates extracted features from satellite data. Forecasting horizons range from t+1 to t+4, corresponding to one to four weeks ahead. 

A different model was selected for each forecasting horizon, as this greatly improved the model performance metrics in comparison to single models with four outputs. LSTM networks were selected more often as the best model, i.e., on 32 occasions, while MLP and CNN were selected 15 and 17 times, respectively, the latter showing a larger expression in lagoon areas (RIAV areas). 

\begin{table}[htbp]
\caption{Summary of the metrics obtained for all areas (best values between conditions highlighted). A different model was selected for each forecasting horizon.}
\label{table:results}
\centering
\resizebox{\textwidth}{!}{\begin{tabular}{@{}cllllllllll@{}}
\hline
\multirow{2}{*}{\textbf{Area}} &
  \multicolumn{1}{c}{\multirow{2}{*}{\textbf{Metrics}}} &
  \multicolumn{4}{c}{\textbf{Univariate}} &
   &
  \multicolumn{4}{c}{\textbf{Multivariate}} \\ \cmidrule(lr){3-6} \cmidrule(l){8-11} 
 &
  \multicolumn{1}{c}{} &
  \multicolumn{1}{c}{\textbf{t+1}} &
  \multicolumn{1}{c}{\textbf{t+2}} &
  \multicolumn{1}{c}{\textbf{t+3}} &
  \multicolumn{1}{c}{\textbf{t+4}} &
   &
  \multicolumn{1}{c}{\textbf{t+1}} &
  \multicolumn{1}{c}{\textbf{t+2}} &
  \multicolumn{1}{c}{\textbf{t+3}} &
  \multicolumn{1}{c}{\textbf{t+4}} \\ \hline
\multirow{4}{*}{L1}    & mae      & \textbf{46}   & \textbf{80}   & 89            & \textbf{86}  &  & 61            & 86            & \textbf{83}   & 93            \\
                       & rmse     & \textbf{79}   & \textbf{112}  & \textbf{126}  & \textbf{128} &  & 92            & 131           & 133           & 144           \\
                       & accuracy & \textbf{0.89} & 0.75          & \textbf{0.74} & 0.72         &  & 0.85          & \textbf{0.77} & 0.72          & \textbf{0.74} \\
                       & model    & LSTM          & MLP           & LSTM          & MLP          &  & LSTM          & MLP           & MLP           & CNN           \\ \hline
\multirow{4}{*}{L2}    & mae      & \textbf{112}  & \textbf{142}  & \textbf{155}  & 165          &  & 117           & 154           & 161           & 164           \\
                       & rmse     & \textbf{154}  & \textbf{183}  & 219           & \textbf{233} &  & 156           & 207           & \textbf{217}  & 235           \\
                       & accuracy & \textbf{0.77} & \textbf{0.66} & \textbf{0.60}  & 0.58         &  & 0.68          & 0.62          & 0.49          & 0.57          \\
                       & model    & LSTM          & LSTM          & LSTM          & LSTM         &  & LSTM          & LSTM          & LSTM          & CNN           \\ \hline
\multirow{4}{*}{L5B}   & mae      & 73            & 82            & \textbf{92}   & 106          &  & \textbf{71}   & \textbf{72}   & 97            & 107           \\
                       & rmse     & 110           & 115           & \textbf{133}  & \textbf{151} &  & \textbf{105}  & \textbf{99}   & 147           & 154           \\
                       & accuracy & 0.92          & 0.94          & \textbf{0.94} & 0.87         &  & \textbf{0.96} & \textbf{0.96} & 0.92          & 0.87          \\
                       & model    & LSTM          & MLP           & CNN           & CNN          &  & MLP           & LSTM          & CNN           & MLP           \\ \hline
\multirow{4}{*}{L6}    & mae      & 45            & 47            & 47            & 47           &  & 45            & \textbf{45}   & \textbf{44}   & 46            \\
                       & rmse     & 72            & 79            & 80            & 82           &  & 73            & \textbf{71}   & \textbf{73}   & \textbf{76}   \\
                       & accuracy & 0.87          & 0.87          & 0.87          & 0.87         &  & 0.87          & 0.87          & 0.87          & 0.87          \\
                       & model    & MLP           & LSTM          & LSTM          & LSTM         &  & LSTM          & LSTM          & LSTM          & LSTM          \\ \hline
\multirow{4}{*}{RIAV1} & mae      & \textbf{53}   & 89            & 112           & 136          &  & 67            & \textbf{86}   & \textbf{104}  & \textbf{93}   \\
                       & rmse     & 107           & 127           & 147           & 172          &  & \textbf{101}  & \textbf{120}  & \textbf{138}  & \textbf{128}  \\
                       & accuracy & \textbf{0.91} & 0.83          & 0.79          & 0.75         &  & 0.89          & \textbf{0.85} & \textbf{0.87} & \textbf{0.77} \\
                       & model    & LSTM          & CNN           & CNN           & CNN          &  & MLP           & CNN           & CNN           & LSTM          \\ \hline
\multirow{4}{*}{RIAV2} & mae      & 97            & 119           & 136           & 142          &  & \textbf{94}   & \textbf{110}  & \textbf{123}  & \textbf{140}  \\
                       & rmse     & 147           & 160           & 178           & 186          &  & \textbf{138}  & \textbf{148}  & \textbf{163}  & 185           \\
                       & accuracy & \textbf{0.77} & 0.74          & 0.75          & 0.74         &  & 0.75          & \textbf{0.81} & 0.74          & 0.74          \\
                       & model    & LSTM          & CNN           & MLP           & CNN          &  & LSTM          & CNN           & MLP           & CNN           \\ \hline
\multirow{4}{*}{RIAV3} & mae      & 84            & 103           & 123           & 130          &  & \textbf{82}   & 104           & \textbf{116}  & \textbf{115}  \\
                       & rmse     & 141           & 175           & 203           & 213          &  & \textbf{139}  & \textbf{168}  & \textbf{194}  & \textbf{189}  \\
                       & accuracy & 0.87          & \textbf{0.83} & 0.79          & 0.77         &  & \textbf{0.89} & 0.81          & 0.79          & \textbf{0.81} \\
                       & model    & LSTM          & CNN           & CNN           & LSTM         &  & LSTM          & LSTM          & MLP           & LSTM          \\ \hline
\multirow{4}{*}{RIAV4} & mae      & \textbf{81}   & 102           & 113           & 113          &  & 90            & \textbf{100}  & \textbf{101}  & \textbf{98}   \\
                       & rmse     & 162           & 151           & 168           & 178          &  & \textbf{142}  & 152           & \textbf{149}  & \textbf{144}  \\
                       & accuracy & \textbf{0.83} & 0.79          & 0.75          & 0.75         &  & 0.79          & 0.79          & \textbf{0.79} & \textbf{0.81} \\
                       & model    & LSTM          & LSTM          & MLP           & CNN          &  & LSTM          & MLP           & MLP           & LSTM          \\ \hline
\end{tabular}}
\end{table}

The prediction errors tend to increase week by week while the accuracy decreases, as expected, with the exception of the L6 area where there are few cases of contamination resulting in similar performances for all weeks.
In areas L1 and L2, the univariate condition outperforms the multivariate in most metrics. In L5B, multivariate performs better on the first two forecasting horizons, but worse on t+3 and similarly on t+4. Results are similar on L6, with a slight advantage for the multivariate case. In the other four areas (RIAV1 to RIAV4) the multivariate case performs best for most metrics and forecasting horizons, except for t+1 on RIAV1 and RIAV4, where there is a loss in MAE. This shows that some areas benefit from the integration of satellite data, improving the models’ predictions.

\begin{figure}[H]
  \centering
   \includegraphics[width=0.7\textwidth]{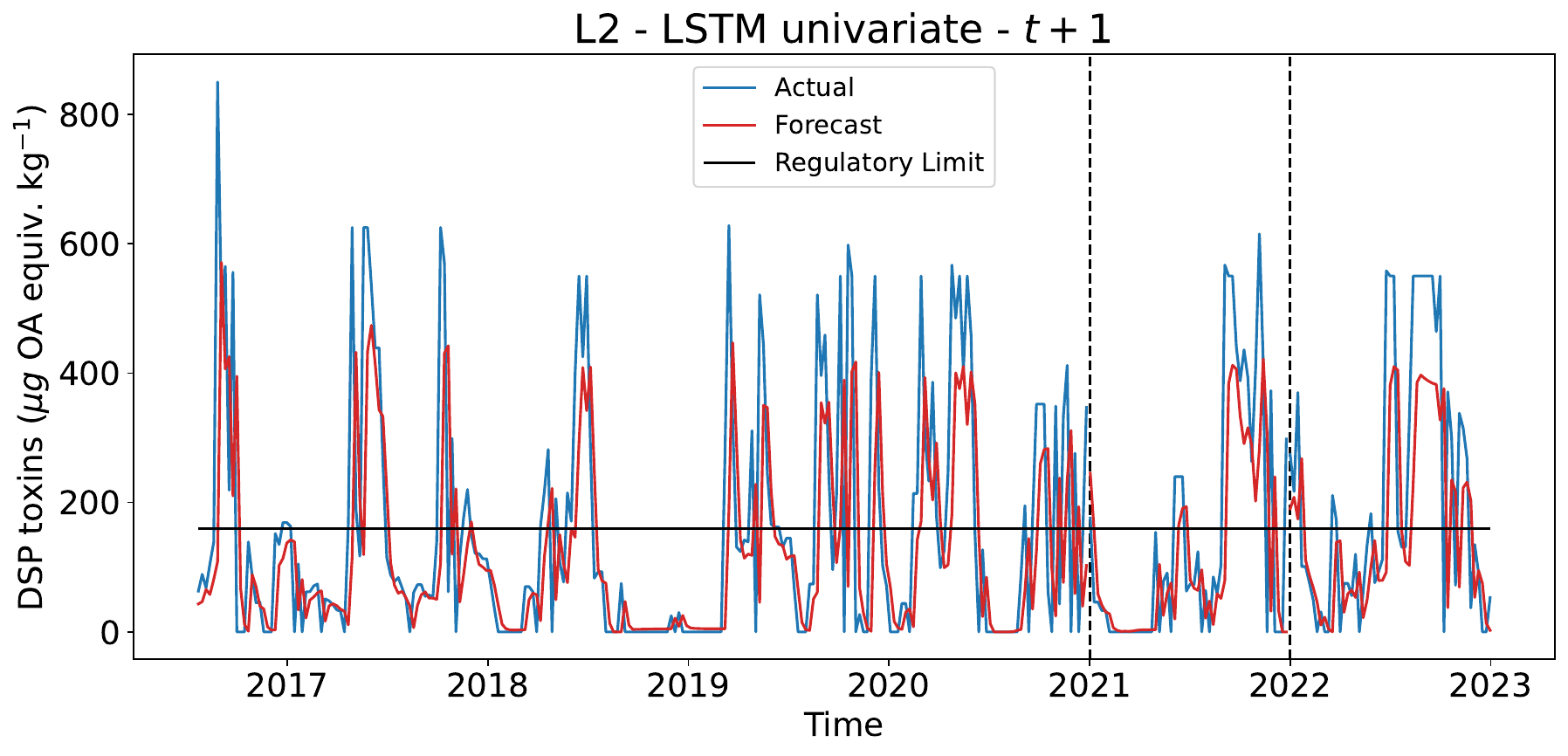}
   \includegraphics[width=0.7\textwidth]{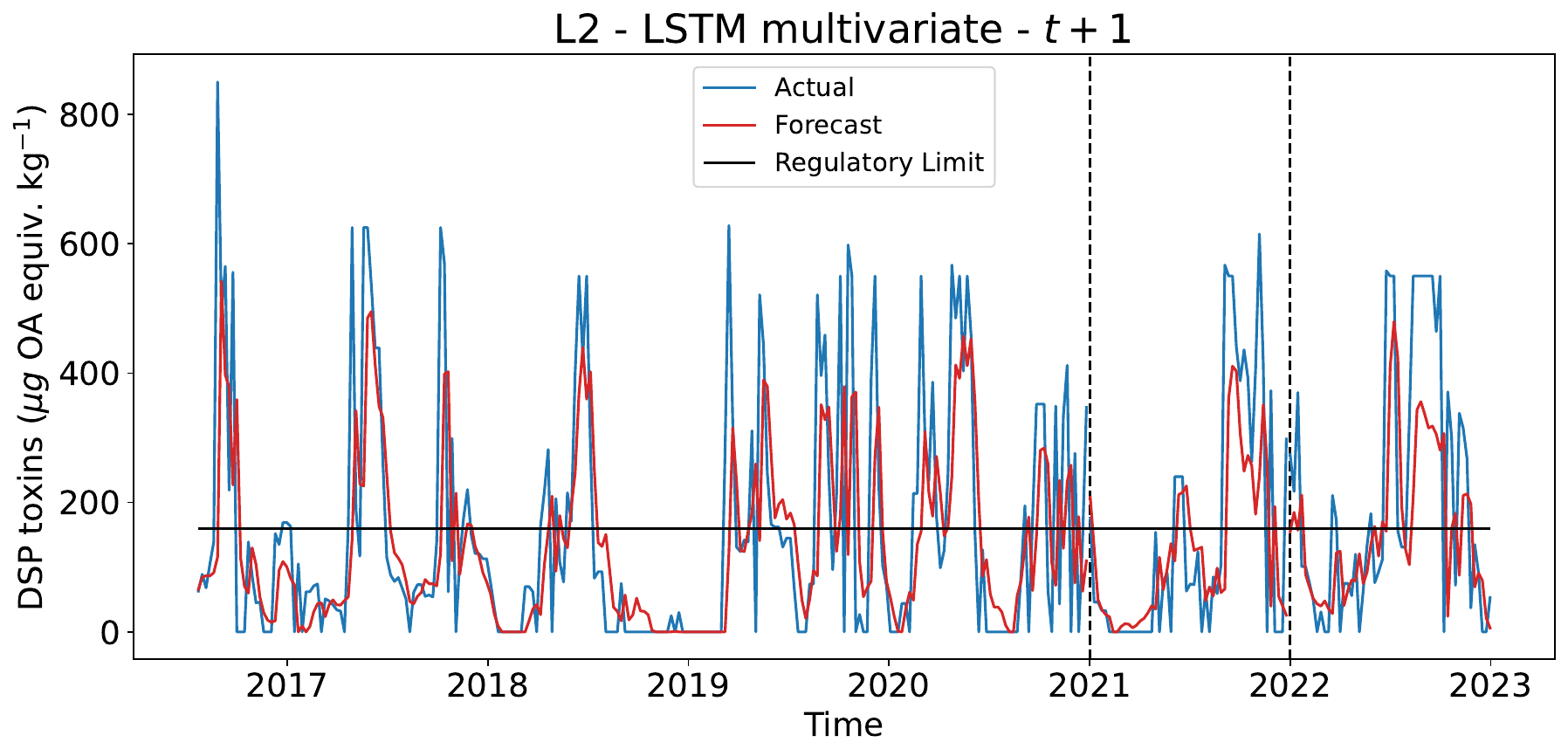}
  \caption{Contamination predictions obtained in area L2 for one week forecasting (t+1). The black line represents the legal limit for contamination (160 $\mu$g OA equiv. $kg^{-1}$), and the dashed lines represent the data split into training, validation, and test sets.}
  \label{fig:graph_results_L2}
\end{figure}

\begin{figure}
\centering
   \includegraphics[width=0.7\textwidth]{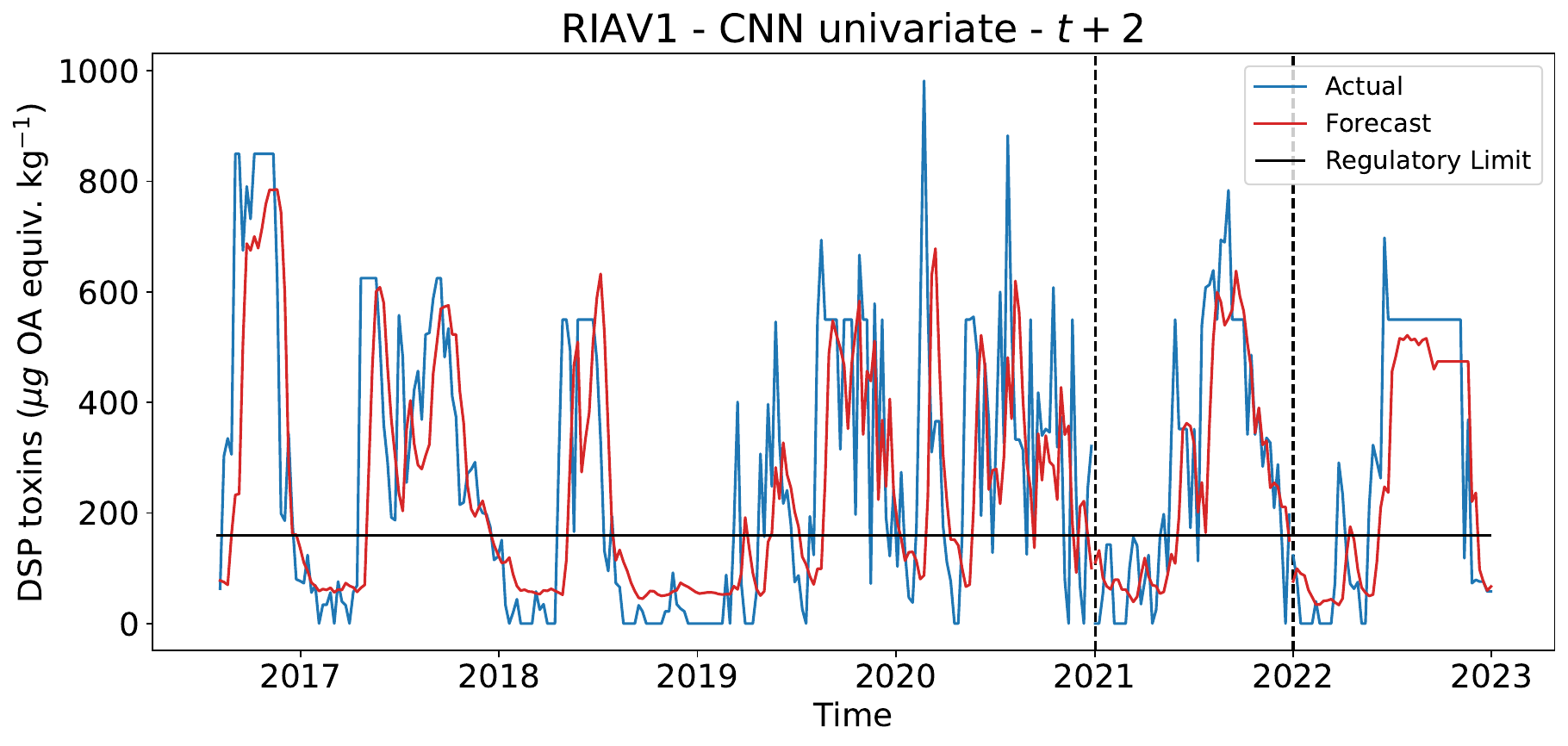}
   \includegraphics[width=0.7\textwidth]{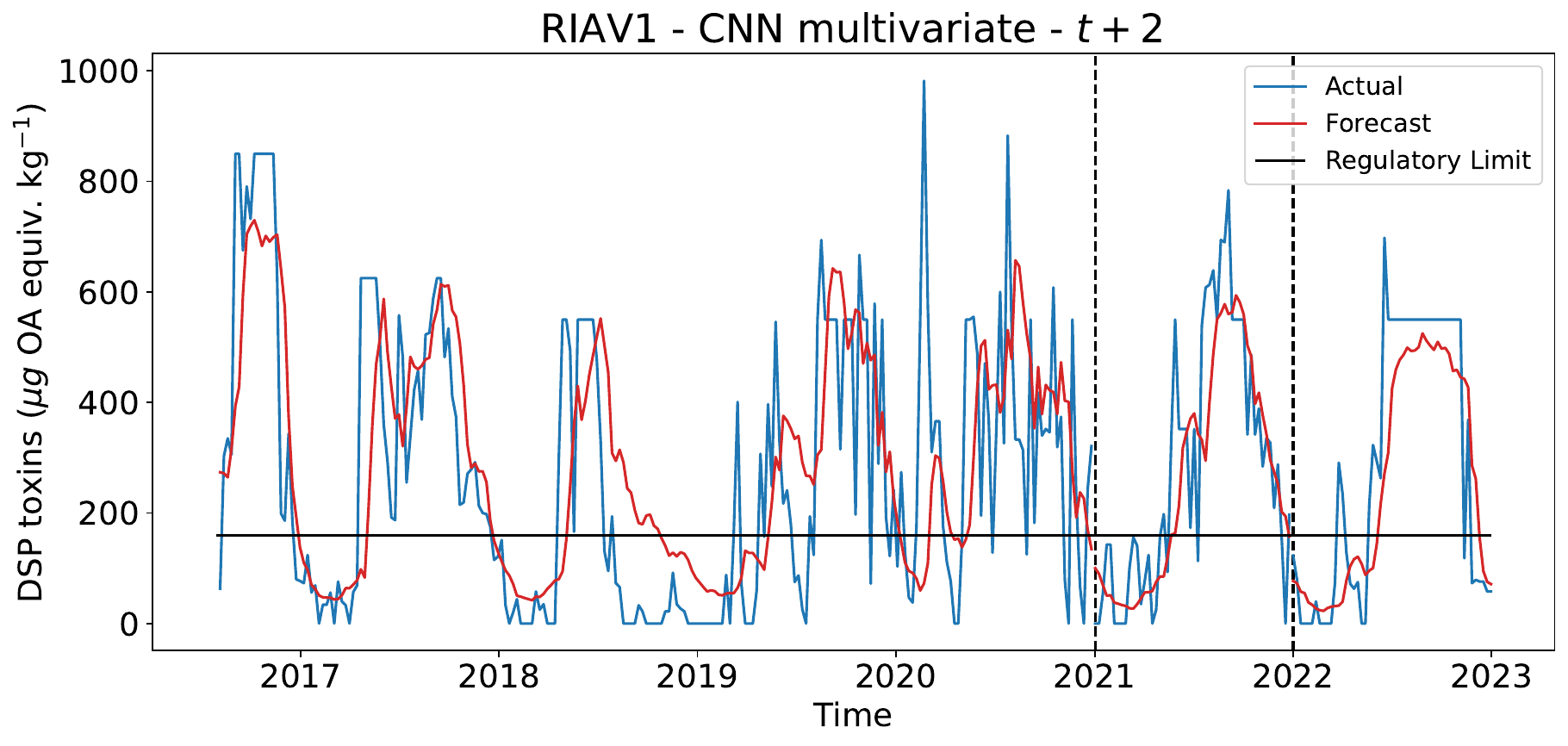}
  \caption{Contamination predictions obtained in area RIAV1 for two week forecasting (t+2). The black line represents the legal limit for contamination (160 $\mu$g OA equiv. $kg^{-1}$), and the dashed lines represent the data split into training, validation, and test sets.}
  \label{fig:graph_results_RIAV1}
\end{figure}

Figures \ref{fig:graph_results_L2} and \ref{fig:graph_results_RIAV1} provide graphical representations of the predictions obtained by the best models in two different regions, one oceanic (L2) and other of a lagoon area (RIAV1). In the first case, for which the inclusion of satellite information did not result in an increase in the predictive ability of the model (Table \ref{table:results}), an increase in the model errors can be observed when this type of information is considered, especially for the contamination cases below the legal safety limit (160 $\mu$g OA equiv. $kg^{-1}$). Even though we use several years of information since sampling occurs, at best, once per week, only hundreds of data points are available to train each model. So, in order to improve predictions outside the training set, the features extracted from the satellite must be sufficiently informative to compensate for the increase in overfitting due to using more attributes. Despite these limitations, this was observed in many cases with evident forecasting improvements.

For the lagoon case (RIAV1), for which an increase in the model prediction performance was obtained when considering satellite information, the metrics are mostly increased for all forecasting horizons, especially for long-term predictions, suggesting a two-week horizon (at least) is informative of a future contamination event. Despite the expected increase in the model prediction errors, such a model might represent a long-term warning to shellfish farmers in this area.

Regarding the accuracy of classification, i.e., below or above the regulatory limit, the results are presented in the form of confusion matrices (Figures \ref{fig:cm_results_L2} and \ref{fig:cm_results_RIAV1}). The accuracy obtained is generally high, indicating an overall good classification. A decrease in accuracy from the univariate to the multivariate case for t+1 forecastings can be observed for the L2 area, whereas an increase in accuracy for t+2 and longer forecasting horizons was obtained for the RIAV1 area (Table \ref{table:results}).

\section{Conclusions}\label{sec4}

{The main objective of this work was to test an approach for extracting useful features from satellite data and evaluate the impact of its integration on forecasting models for predicting toxin concentrations in shellfish. We hypothesize that information gathered from satellite images of a given shellfish production area, particularly those covering the oceanographic neighbouring area, might contain relevant information regarding the marine conditions that lead to contamination events. We investigated how informative these features were by including them as additional features in several ANN-based time-series forecasting models to predict biotoxin contamination in shellfish. 

\begin{figure}
  \centering
  \includegraphics[width=0.35\textwidth]{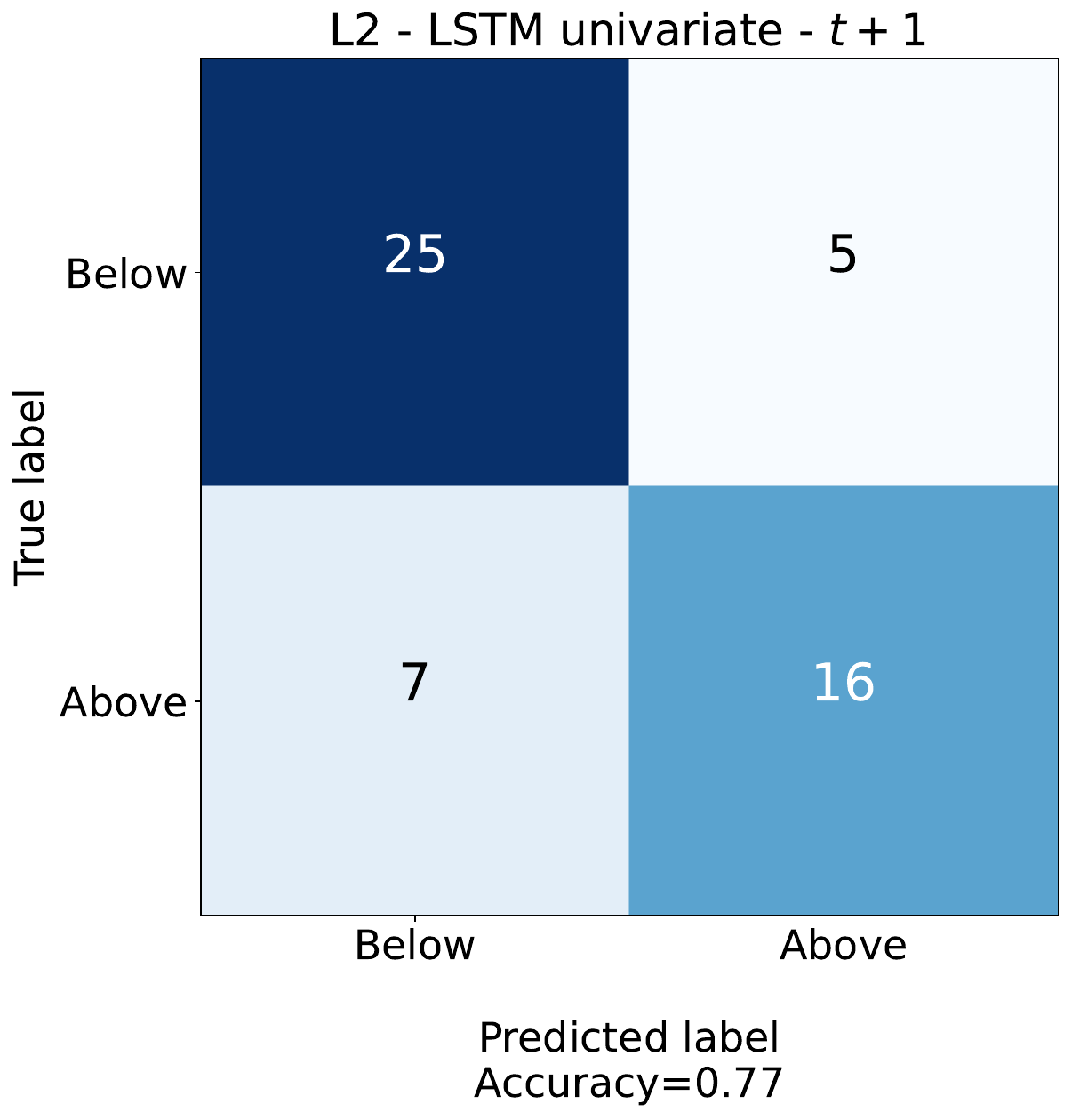}
   \includegraphics[width=0.35\textwidth]{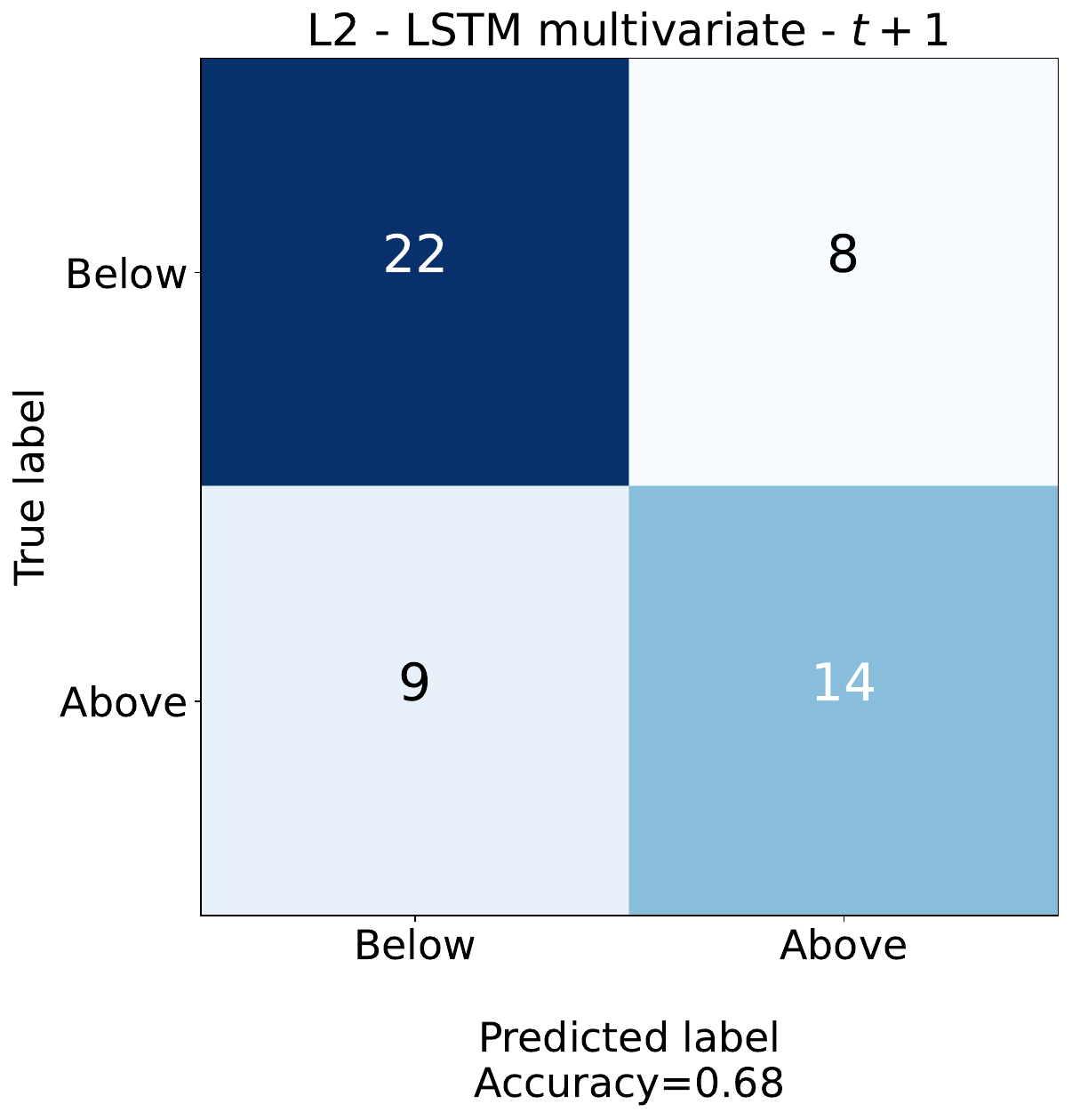}
\caption{Confusion matrices obtained in area L2 for one week forecasting (t+1).}
\label{fig:cm_results_L2}
\end{figure}

\begin{figure}
\centering
\includegraphics[width=0.35\textwidth]{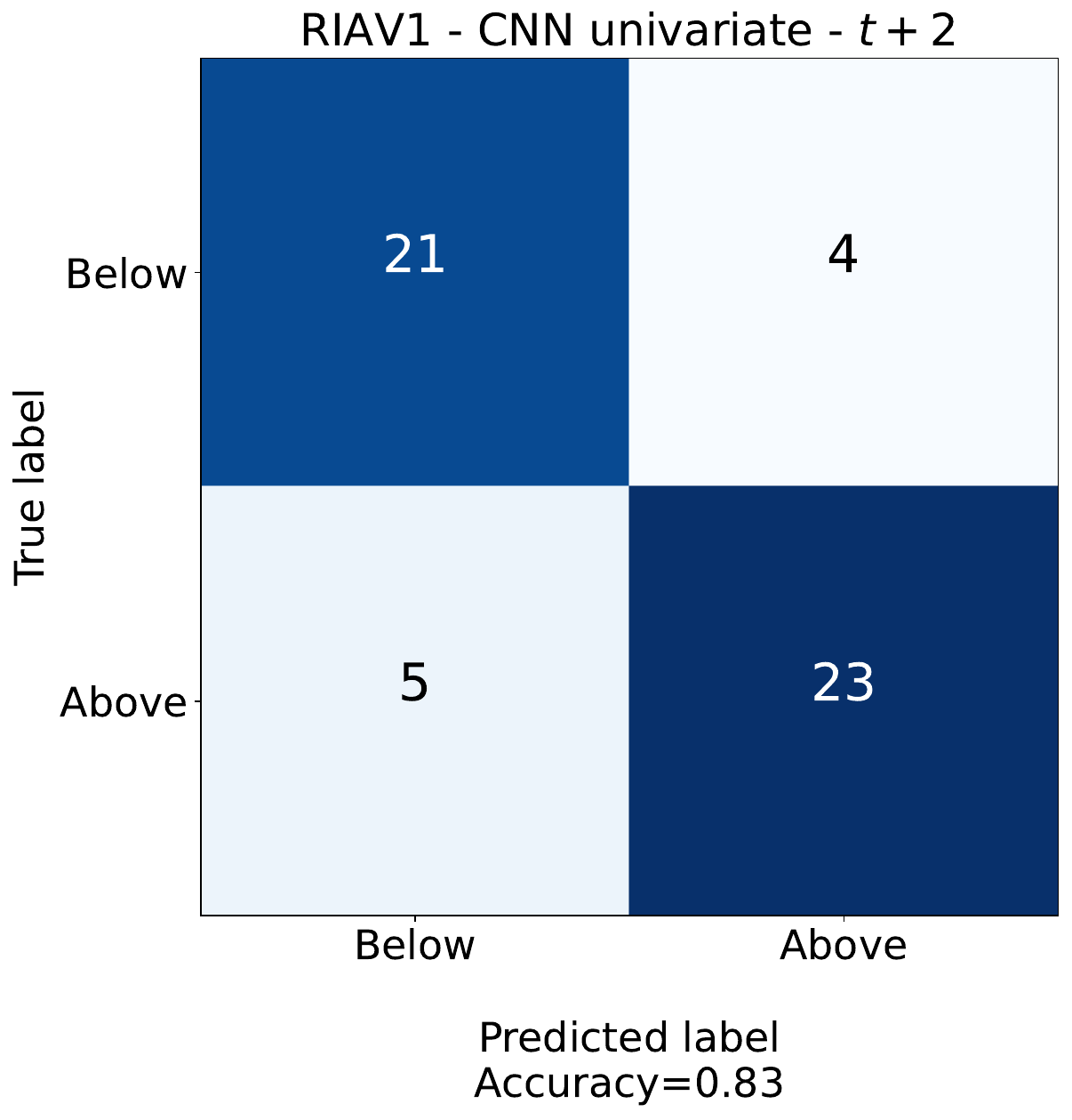}
\includegraphics[width=0.35\textwidth]{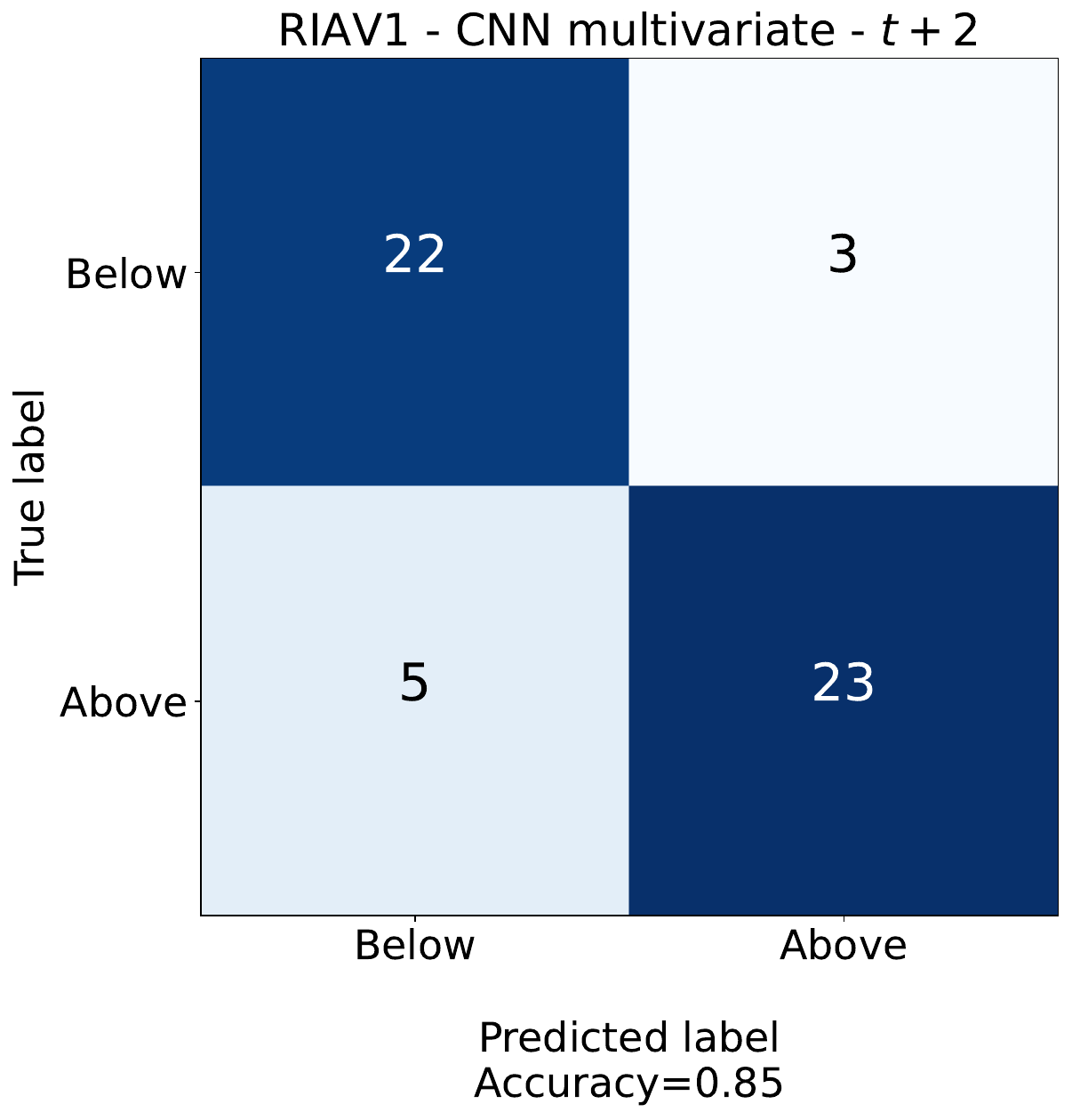}
\caption{Confusion matrices obtained in area RIAV1 for two week forecasting (t+2).}
\label{fig:cm_results_RIAV1}
\end{figure}

The results obtained by the proposed approach show that including satellite data features improves the prediction of contamination values in most areas, particularly for the 2-week and longer forecasting horizons in the lagoon shellfish production areas (RIAV), and for 1-week and 2-week horizons in the L5B area. These findings support the use of autoencoders for feature extraction on satellite data, suggesting that it is possible to include information from a high-dimension data source without losing the ability to generalize outside the training set. Improvements were obtained for most areas studied by integrating only four features, which fosters further investigation on the usefulness of remotely sensed data on shellfish contamination forecasting. Further work might also include refining the selection of image locations, as well as experimenting with different image sizes (current images considered 20$\times$20km) and evaluating its impact on the quality of the information generated. Testing more spectral bands may also be a path for further improvement.

To the best of our knowledge, this work is the first to report improvements in the use of satellite imagery for directly forecasting shellfish contamination. This contribution may pave the way for the development of more robust and accurate predictive systems in the future, anticipating an effective impact on shellfish production management.

\section*{Acknowledgements}

\noindent This work was supported by national funds through the Fundação para a Ciência e a Tecnologia (FCT, I.P.) through the project MATISSE: A Machine Learning-Based Forecasting Systems for Shellfish (DSAIPA/DS/0026/2019), projects UIDB/04516/2020\\ (NOVA LINCS), UIDB/00297/2020 and UIDP/00297/2020 (NOVA Math), UIDB/00667\\ /2020 and UIDP/00667/2020 (UNIDEMI), and also CEECINST/00042/2021.

\bibliographystyle{elsarticle-num} 
\bibliography{Tavares_et_al_arXiv}

\end{document}